\documentclass[10pt,twocolumn]{article}

\usepackage{cvpr}
\usepackage{times}
\usepackage{epsfig}
\usepackage{graphicx}
\usepackage{amsmath}
\usepackage{amssymb}
\usepackage{algorithm}
\usepackage{algorithmic}
\usepackage{multirow}
\usepackage{array}
\usepackage{booktabs}
\usepackage{makecell}
\usepackage{float}
\usepackage{diagbox}
\usepackage{changepage}
\usepackage{cuted} 
\usepackage{xcolor}
\usepackage{caption}
\usepackage{subcaption}
\usepackage{dsfont}

\usepackage[pagebackref=true,breaklinks=true,letterpaper=true,colorlinks,bookmarks=false]{hyperref}

\usepackage[capitalize]{cleveref}
\crefname{section}{Sec.}{Secs.}
\Crefname{section}{Section}{Sections}
\Crefname{table}{Table}{Tables}
\crefname{table}{Table}{Tables}

\begin{document}

\title{BoxVIS: Video Instance Segmentation with Box Annotations}

\author{Minghan Li, Lei Zhang\thanks{Corresponding author.} \\
The Hong Kong Polytechnic University \\
{\tt\small liminghan0330@gmail.com}, 
{\tt\small cslzhang@comp.polyu.edu.hk}
}

\maketitle

\begin{abstract}
It is expensive and labour-extensive to label the pixel-wise object masks in a video. As a result, the amount of pixel-wise annotations in existing video instance segmentation (VIS) datasets is small, limiting the generalization capability of trained VIS models. 
An alternative but much cheaper solution is to use bounding boxes to label instances in videos. Inspired by the recent success of box-supervised image instance segmentation, we adapt the state-of-the-art pixel-supervised VIS models to a box-supervised VIS (BoxVIS) baseline, and observe slight performance degradation. 
We consequently propose to improve the BoxVIS performance from two aspects. First, we propose a box-center guided spatial-temporal pairwise affinity (STPA) loss to predict instance masks for better spatial and temporal consistency. 
Second, we collect a larger scale box-annotated VIS dataset (BVISD) by consolidating the videos from current VIS benchmarks and converting images from the COCO dataset to short pseudo video clips. 
With the proposed BVISD and the STPA loss, our trained BoxVIS model achieves 43.2\% and 29.0\% mask AP on the YouTube-VIS 2021 and OVIS valid sets, respectively.
It exhibits comparable instance mask prediction performance and better generalization ability than state-of-the-art pixel-supervised VIS models by using only 16\% of their annotation time and cost. Codes and data can be found at \url{https://github.com/MinghanLi/BoxVIS} .
   
\end{abstract}
\section{Introduction}
Video instance segmentation (VIS) aims to predict the pixel-wise masks and categories of instances over the input video. The recently proposed VIS methods \cite{wang2020vistr,hwang2021video,seqformer,heo2022vita,IDOL, huang2022minvis} have made great progress on the YouTube-VIS \cite{yang2019video} and OVIS \cite{qi2021occluded} benchmark datasets. Generally speaking, these methods first partition the whole video into individual frames or short clips to extract per-frame or per-clip features and predict instance masks, and then associate the predicted masks across frames/clips based on the embedding similarity of instances. However, all the current VIS methods require pixel-wise annotations, where the pixels belonging to the target instances are labeled as 1 and others as 0, to train the VIS model for pixel-wise mask prediction.

\begin{table}
\centering
\begin{tabular}{p{0.12\textwidth}p{0.045\textwidth}<{\centering}p{0.05\textwidth}<{\centering}p{0.075\textwidth}p{0.065\textwidth}p{0.05\textwidth}<{\centering}}
\Xhline{0.8pt}
{\normalsize Methods} & {\normalsize Superv.}  & { Annot.} & {\normalsize YTVIS21} &  {\normalsize OVIS} \\ 
\Xhline{0.8pt}
{\normalsize M2F-VIS}        & Pixel & { 485d} & 43.2 & {24.5 } \\  
{\normalsize M2F-VIS-box}    & Box   & {  43d}  & 41.2$^{{ -2.0}}$ & 23.9$^{{-0.6}}$ \\  
{\normalsize BoxVIS (ours)}   & Box   & {  80d}  & 43.2$^{{ +0.0}}$  & 29.0$^{{+4.5}}$ \\
\Xhline{0.8pt}
\end{tabular}
\vspace{-2mm}
\caption{ Performance (Mask AP) comparison of pixel-supervised Mask2Former-VIS (M2F-VIS) \cite{cheng2021mask2former-video}, its box-supervised counterpart M2F-VIS-box, and our BoxVIS on YTVIS21 and OVIS valid sets.}\label{tab:min_boxvis}
\vspace{-2mm}
\end{table}

Compared with image instance segmentation (IIS), VIS requires a larger number of annotated training data due to the existence of object motion and complex trajectories in videos. Unfortunately, it is an expensive and labour-extensive task to label the pixel-wise object masks in a video. As a result, the amount of pixel-wise annotations in existing VIS datasets is small. For example, there are only 8k and 5k labeled instances in YouTube-VIS 2021 (YTVIS21) and OVIS, respectively, which limits the generalization capabilities of trained VIS models. However, it has been demonstrated \cite{papadopoulos2017extreme,lin2014microsoft} that labeling the bounding box of an object takes only 8.8\% (7s vs. 79.2s) the time of labeling its polygon-based mask in COCO \cite{lin2014microsoft}. 
Actually, a few box-supervised IIS methods \cite{lan2021discobox,tian2020boxinst,li2022boxlevelset,cheng2022boxteacher,lan2023MAutoLabeler} have been recently developed and achieved competitive performance with pixel-supervised IIS methods.
Therefore, one natural and interesting question is: can we use bounding boxes to label instances to train the VIS models?  

To validate the feasibility of VIS with only box annotations, we adapt the state-of-the-art pixel-supervised VIS model to a box-supervised VIS (BoxVIS) baseline. Specifically, inspired by the box-supervised IIS methods BoxInst \cite{tian2020boxinst} and BoxTeacher \cite{cheng2022boxteacher}, we introduce the box-supervised segmentation loss terms into the pixel-supervised VIS method Mask2Former-VIS (M2F-VIS) \cite{cheng2021mask2former-video}, and replace the ground-truth masks with the produced pseudo masks to supervise the model learning. 
(Please refer to Section \ref{sec:boxvis_baseline} for details.)  
The adapted BoxVIS model from M2F-VIS is termed as M2F-VIS-box. As shown in \cref{tab:min_boxvis}, compared with M2F-VIS, M2F-VIS-box drops only 2.0\% and 0.6\% mask AP on YTVIS21 and OVIS valid sets, respectively. We think there are two main reasons for the slight performance degradation. 
First, all current VIS models have been pre-trained on the COCO dataset before fine-tuned on the VIS datasets. There is a large overlap of categories between COCO and VIS datasets so that the pre-training enables a good baseline VIS model. Second, when fine-tuning on videos, ground-truth boxes and pseudo masks actually provide comparable supervision to pixel-wise annotation to learn object motion and appearance changes. 

Nonetheless, there are two challenging issues for BoxVIS.
First, without pixel-wise labeling, the box-level annotation cannot explicitly tell the segmenter the precise object boundary and the temporal association of objects. It is essential to investigate how to make the VIS models have good object spatial-temporal consistency with only box annotations.
Second, objects in videos often have significant position shifts, appearance changes, heavy occlusion, uncommon camera-to-object views, \etc. To deal with the diverse variations of videos, a larger scale box-annotated dataset is required to train robust BoxVIS models. 

With the above considerations, we propose to improve the BoxVIS performance from two aspects: modeling the object spatial-temporal consistency and increasing the amount of box-annotated video clips. 
First, we propose a box-center guided spatial-temporal pairwise affinity (STPA) loss to predict instance masks with better spatial and temporal consistency.
Second, we collect a larger box-annotated VIS dataset (BVISD) by consolidating the videos from current VIS benchmarks and converting some images from COCO to short pseudo video clips. 
The trained BoxVIS model with the proposed STPA loss and BVISD  demonstrates promising instance segmentation results, achieving 43.2\% and 29.0\% mask AP on YouTube-VIS 2021 and OVIS valid sets, respectively. It obtains comparable or even better generalization performance than the pixel-supervised VIS competitors by using only 16\% of their annotation time and cost, exhibiting great potentials.

\section{Related Work}

\subsection{Pixel-supervised VIS}
There are two major VIS benchmarks, the YTVIS series \cite{yang2019video} and OVIS \cite{qi2021occluded}, which have very different video types in terms of object motions and scenes. The YTVIS series focus mainly on segmenting sparse objects in shorter videos, while the OVIS aims to segment crowded instances with occlusions in longer videos. Based on these facts, we categorize the pixel-supervised VIS models into YTVIS-oriented ones and OVIS-oriented ones.

\textbf{YTVIS-oriented VIS models.}
By introducing a tracker into the representative IIS methods \cite{he2017mask,bolya2019yolact,tian2020conditional}, the early proposed VIS methods \cite{yang2019video, cao2020sipmask,Li_2021_CVPR,Athar_Mahadevan20stemseg,liu2021sg,yang2021crossover,QueryInst,ke2021pcan} have achieved decent performance on YTVIS series. 
The frame-to-frame trackers \cite{yang2019video,cao2020sipmask,Li_2021_CVPR,QueryInst} integrate the clues such as category score, box/mask IoU and instance embedding similarity. 
The clip-to-clip trackers \cite{Li_2021_CVPR,Athar_Mahadevan20stemseg,bertasius2020classifying,lin2021video, qi2021occluded,wang2020vistr,seqformer,hwang2021video} propagate the predicted instance masks from a key frame to other frames\cite{bertasius2020classifying,dai2017deformable,lin2021video,Li_2021_CVPR,qi2021occluded,wang2021end}.
Recently, query-based \cite{carion2020end} VIS methods \cite{wang2020vistr,hwang2021video,seqformer,wu2022trackletquery} have achieved impressive progress. VisTR \cite{wang2020vistr} views the VIS task as an end-to-end parallel sequence prediction problem. 
To reduce the storing memory of spatial-temporal features, IFC\cite{hwang2021video} transfers inter-frame information via efficient memory tokens, and SeqFormer \cite{seqformer} locates an instance in each frame and aggregates them to predict video-level instances. 

\textbf{OVIS-oriented VIS models.}
The aforementioned YTVIS-oriented VIS models often fail to handle the challenging long videos with crowded and similar-looking objects in OVIS dataset, resulting in significant performance degradation. Inspired by contrastive learning \cite{chen2020simplectt, pang2021quasi,khosla2020supervisedctt,wang2021densectt}, IDOL \cite{IDOL} learns discriminative embeddings for multiple object tracking frame by frame.
Mask2Former \cite{cheng2021mask2former} achieves impressive performance on IIS tasks by calculating attention only in the region of objects.
M2F-VIS \cite{cheng2021mask2former-video} and MinVIS \cite{huang2022minvis} extend Mask2Former to VIS task, where they respectively take per-frame and per-clip inputs.
VITA \cite{heo2022vita} integrates object embeddings of all frames in the video to produce video-level instance masks.

\textbf{Remarks.} Though the pixel-supervised VIS methods have achieved much progress, their generalization capability is limited. For example, the VIS models trained on YTVIS21 often fail to handle the challenging videos in OVIS, due to the limited number of type of videos in each dataset. However, it is labour-extensive to label the pixel-wise masks in videos. Inspired by the success of box-supervised IIS methods, we explore VIS task with only box annotations in this paper.

\begin{figure*}[!t]
\begin{center}
    \includegraphics[width=0.98\linewidth]{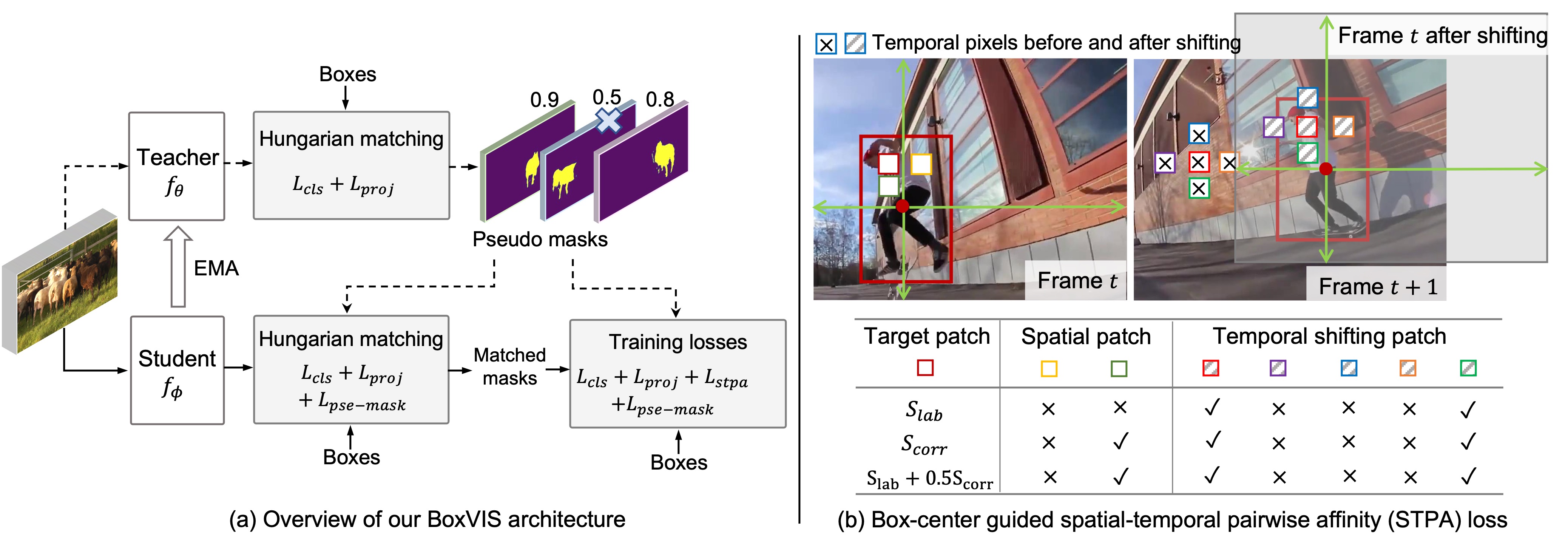}
\end{center}
\vspace{-5mm}
\caption{(a) Architecture of our proposed BoxVIS, where the Teacher and Student Nets follow the Mask2Former-VIS \cite{cheng2021mask2former-video} framework. The Teacher Net produces high-quality object masks, which are assigned as the pseudo instance masks of the ground-truth bounding boxes by Hungarian matching. 
The predicted masks from the Student Net are matched with the ground-truth boxes and the generated pseudo masks by taking the assigned pseudo masks into account.
(b) Schematic diagram of the proposed spatial-temporal pairwise affinity (STPA) loss, which uses box-center guided shifting to generate temporally paired pixels and employs the color similarity $S_{lab}$ and patch correlation $S_{corr}$ to compute pairwise affinity $S_e=S_{lab}+0.5S_{corr}$. `$\times$' and `$\checkmark$' mean that the value is below or above the threshold.}
\label{fig:boxvis_baseline}
\end{figure*}
\subsection{Weakly-supervised Segmentation}

\textbf{Box-supervised IIS.}
A few box-supervised IIS methods have been proposed. BoxCaseg \cite{wang2021boxcaseg} leverages a saliency model to generate pseudo object masks. BoxInst  \cite{tian2020boxinst} achieves impressive performance by proposing a projection loss and a pairwise affinity loss for box-supervision. BoxLevelSet \cite{li2022boxlevelset} utilizes the traditional level set evolution to predict the object boundaries and instance masks. DiscoBox \cite{lan2021discobox} and BoxTeacher \cite{cheng2022boxteacher} employ an exponential moving average (EMA) teacher to produce high-quality pseudo masks and introduce pseudo pixel-wise mask supervision, bringing significant performance improvement. 

\textbf{Weakly-supervised Video Segmentation.}
For the video object segmentation (VOS) task, BoxVOS \cite{hannan2022box} and QMRA \cite{lin2021query} utilize the motion map and feature aggregation from consecutive frames to segment objects in videos with only box supervision. 
For the VIS task, FlowIRM \cite{liu2021weakly} presents a class-supervised VIS baseline with relatively low performance. In this work, we propose the first box-supervised VIS framework and demonstrate its effectiveness. 

\section{Methodology}

We first extend the pixel-supervised VIS method to a box-supervised VIS (BoxVIS) baseline in \cref{sec:boxvis_baseline}, then propose a box-center guided spatial-temporal pairwise affinity (STPA) loss and a large scale box-annotated VIS dataset (BVISD) in \cref{sec:st_pair} and \cref{sec:bvisd}, respectively.

\subsection{BoxVIS baseline}\label{sec:boxvis_baseline}
In this subsection, we first briefly introduce the box-supervised segmentation loss \cite{tian2020boxinst} and the pseudo mask supervision loss with the high-quality pseudo masks \cite{lan2021discobox,cheng2022boxteacher}, then extend the per-clip input based M2F-VIS model \cite{cheng2021mask2former-video} to its box-supervised counterpart, namely M2F-VIS-box, as a BoxVIS baseline.

\textbf{Box-supervised segmentation loss \cite{tian2020boxinst}} consists of the projection loss and the pairwise affinity loss. Projection loss supervises  the  horizontal  and  vertical  projections of predicted masks using the ground-truth bounding boxes via Dice loss \cite{dice1945dice, milletari2016dice_vnet}, termed as ${L}_{proj}$.
Pairwise affinity loss is designed to supervise pixel-wise mask prediction without pixel-wise annotation. 
For two pixels, if their color similarity is grater than a threshold, they are assumed to have the same mask label, denoted as $y_e = 1$. Their pairwise mask affinity is $ P(y_e=1) = {M}_i\cdot {M}_j + (1-{M}_i)\cdot(1-{M}_j)$,
where ${M}_i$ and ${M}_j$ represent the predicted masks of the two pixels, respectively.
The pairwise affinity loss is
\begin{equation}\label{eq:pair}
   \vspace{-1mm}
   {L}_{pair}  =  -\frac{1}{N} \sum\limits_{e\in E_{in}} \mathds{1}_{\{S_{lab} \geq \tau_{lab}\}} \log P(y_e = 1).
\end{equation}
where $S_{lab}$ and $\tau_{lab}$ are the color similarity and the color threshold in the LAB space, respectively. $E_{in}$ indicates the set of paired pixels in the inner bounding box, and $N$ is the total number of the pixel pairs with the same labels in $E_{in}$.

\textbf{Pseudo mask supervision loss.} Exponential moving average (EMA) teacher \cite{lan2021discobox,cheng2022boxteacher} can generate high-quality pseudo instance masks of the ground-truth boxes. By replacing the ground-truth masks with the pseudo masks, the pixel-wise mask supervision loss can be extended accordingly to the pseudo mask supervision loss, termed as $L_{pse\text{-}mask}$, which usually consists of the binary cross-entropy (BCE) loss and the Dice loss \cite{dice1945dice, milletari2016dice_vnet}.

\textbf{BoxVIS baseline.} By using the above two box-supervised loss terms and the pseudo mask supervision loss, we extend the pixel-supervised VIS method M2F-VIS \cite{cheng2021mask2former-video} to a BoxVIS baseline, namely M2F-VIS-box.
The overall architecture of the BoxVIS baseline is shown in \cref{fig:boxvis_baseline}(a). 
During training, a short video clip will be fed to the sophisticated Teacher Net and the Student Net to simultaneously predict object masks. 
On one hand, the predicted masks from the Teacher Net will be taken as the pseudo masks of the ground-truth boxes by the Hungarian matching algorithm, whose cost matrix consists of the classification loss and the projection loss. 
On the other hand, the predicted masks from the Student Net will be matched with the ground-truth boxes and the generated pseudo masks, whose matching cost matrix takes the assigned pseudo masks into account. 
Finally, the matched masks from the Student Net are supervised by the ground-truth boxes and the pseudo masks via the box-supervised segmentation loss and the pseudo-mask supervision loss, respectively. 
Note that the parameters of the Teacher Net are progressively updated via EMA.
During inference, the Teacher Net will be discarded, and we only use the predicted masks by the Student Net.

Beside, the low-quality pseudo masks may introduce some incorrect pixel-wise supervision for the pseudo-mask loss, resulting in performance degradation. The confidence scores for estimating the mask quality are computed as the product of the classification score and the projection score. To select high-quality pseudo masks, we adopt a dynamic threshold, which is determined by the ratio of the number of processed iterations to the total number of training iterations: $\epsilon = 1/(1+e^{-2r})$, where $r={\text{iters}}/{\text{total iters}}$.

\subsection{Spatial-temporal Pairwise Affinity Loss} \label{sec:st_pair}
Segmenting instances from videos is more challenging than segmenting objects from individual images, as videos often have unexpected motion blur, uncommon-camera-to-object motion, appearance changes, heavy occlusion and so on.  
The pairwise affinity loss in Eq. (\ref{eq:pair}) only considers the paired pixels within a single frame, which may however fail for BoxVIS to constrain the temporal consistency of segmented objects. 
One naive solution is to directly select the neighbours at the same positions in the next frame to perform temporal pairwise affinity loss. Unfortunately, objects in a video often change their positions due to camera jitters or object motion. The assumption that two adjacent pixels in an image will have similar colors and hence the same mask labels does not hold well for adjacent frames in a video. That is, pixels from two adjacent frames may have similar colors but they can belong to different instances.
To solve this issue, we propose a box-center guided spatial-temporal pairwise affinity (STPA) loss, which uses the box-center guided shifting to generate the temporally paired pixels, and uses the color similarity and the patch feature correlation to stably compute the pairwise affinity.

\textbf{Spatial-temporally paired pixels.}
With an undirected graph, combining a pixel's left, top, right and bottom neighbours is equal to combining its right and bottom neighbours, but the latter is more efficient in training.
For a pixel in a video clip, we consequently only consider its right and bottom neighbours in the current frame, namely the set $E_{s}$ of spatial neighbours. 
We then discuss how to use the ground-truth bounding boxes to identify inter-frame paired pixels. The ground-truth bounding boxes in consecutive frames can indicate coarse inter-frame movements of the objects. 
For an instance that appears in two adjacent frames $t_i$ and $t_j$, we denote by $(t_i, x_i^c, y_i^c)$ and $(t_j, x_j^c, y_j^c)$ its bounding box centers, and by $(dx^c, dy^c) = (x_j^c-x_i^c, y_j^c-y_i^c)$ the location offsets of the two centers.
Then, for any pixel at position $(t_i, x_i, y_i)$ in frame $t_i$, if it locates in the inner bounding box of the instance, we can shift its position to the nearby area of the instance bounding box in frame $t_j$ by $(t_j, x_i+dx^c, y_i+dy^c)$.
Consequently, the set $E_{t}$ of temporal neighbours can be produced by shifting the center pixel and its four spatial neighbours in frame $t_i$ to the corresponding positions in frame $t_j$. 
Overall, for each pixel, we group its seven neighbours as paired pixels to compute the proposed STPA loss. The schematic and the generated spatial-temporal pixels are illustrated  in \cref{fig:boxvis_baseline}(b).

\textbf{Patch feature correlation.}
It is not reliable enough to utilize pixel-to-pixel color similarity to determine whether two pixels have the same labels or not. Therefore, we introduce an extra patch feature correlation to compute stable and reliable temporal pairwise affinity. 
For a pixel at position $(t_i, x_i, y_i)$, we represent the pixels in its centered patch as $(t_i, x_i+o_x, y_i+o_y)$, where the displacements $(o_x, o_y) \in {\normalsize [-k, k]\times [-k, k]}$ and $k$ controls the patch size. The feature correlation of two patches is 
\begin{equation}
 \begin{normalsize}
 {S}_{corr} = \frac{1}{|\Omega|} \sum_{o \in \Omega}\ \frac{<f_{i+o},\ f_{j+o}>}{\parallel f_{i+o} \parallel \cdot \parallel f_{j+o} \parallel }, 
 \end{normalsize}
\end{equation}
where $f_{i+o}$ and $f_{j+o}$ are the feature vectors at positions $(t_i, x_i+o_x, y_i+o_y)$ and $(t_j, x_j+o_x, y_j+o_y)$ in the last pixel encoder layer of Mask2Former framework. $|\Omega|$ is the total number of pixels within the patch $\Omega = {\normalsize [-k, k]\times [-k, k]}$, and $k$ is set as 1 by default.

The overall pairwise affinity is $S_e = S_{lab} + 0.5S_{corr}$.
To remove the paired pixels with low affinity, we introduce a color similarity threshold $\tau_{lab}$ and a correlation threshold $\tau_{corr}$, respectively. The affinity threshold used in our paper is $\tau=\tau_{lab}+0.5\tau_{corr}$. Without otherwise specified, we set $\tau_{lab}=0.3$ and $\tau_{corr}=0.9$ by default, \ie, $\tau=0.75$.
Finally, the STPA loss can be formulated as:
\begin{align}\label{eq:stpair}
\begin{normalsize}
   {L}_{stpa}  =  -\frac{1}{N} \sum\limits_{e\in E_{stin}} \mathds{1}_{\{S_e \geq \tau\}} \log P(y_e = 1).
\end{normalsize}
\end{align}
where $P(y_e=1)$ is the mask pairwise affinity \cite{tian2020boxinst}, and $E_{stin}$ indicates the set of the spatial-temporally paired pixels, among which at least one pixel locates in the inner bounding box of the instance in current frame. $N$ is the total number of the paired pixels with affinity higher than the threshold $S_e$ in the set $E_{stin}$. 

\subsection{Box-annotated VIS dataset (BVISD) } \label{sec:bvisd}
It is very costly to annotate fine-grained masks for instances in videos. 
We propose a larger scale box-annotated VIS dataset (BVISD) by merging the videos from current VIS benchmarks (\ie, YouTube-VIS and OVIS), and converting images from COCO to short pseudo video clips. 
For YouTube-VIS benchmark, we adopt the latest YTVIS21, which contains 2,985 training videos over 40 categories, and the video length is less than 36 frames. OVIS \cite{qi2021occluded} includes 607 training videos over 25 categories, but each frame has crowded objects with different occlusion levels. 

\begin{figure}[t]
     \vspace{-1mm}
     \includegraphics[width=\linewidth]{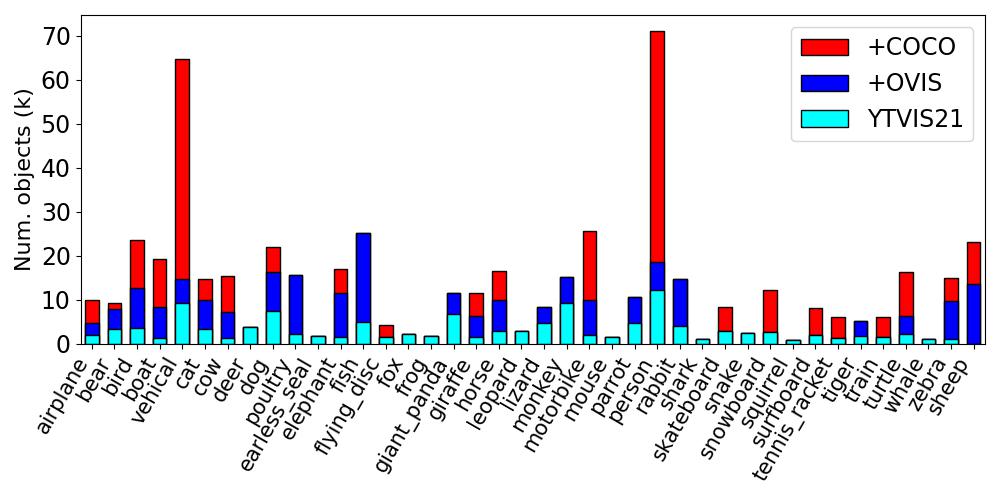}
     \vspace{-8mm}
     \caption{ Number of objects  per category in BVISD. The number of `person' category is divided by 5 for better display.}\label{fig:bvisd_cate}
     \vspace{-2mm}
\end{figure}

There are 25 overlapping categories between COCO and VIS benchmarks, including around 90k images and 450k objects. To augment the single image from COCO to a video clip, we first resize the image by adjusting its short edge in the range of [600, 800], and then randomly crop a region with the short edge within [320, 512]. Finally, the cropped regions are randomly rotated within the degree of [-15, 15]. We repeat the above augmentation process for $T$ times to obtain a pseudo video clip with $T$ frames. During training, the short video clip augmented from COCO will be further resized into the input resolution, \ie 360p. An example of pseudo clip generation is illustrated in the \textbf{supplemental materials}.

One key step to consolidate the above three datasets is how to properly  merge the object categories to avoid semantic conflicts.
Fortunately, the 25 categories of OVIS are mostly contained in the 40 categories of YTVIS21, except for the `sheep' category. Besides, the categories `car' and `truck' in YTVIS21 are merged into a super-category `vehicle' in OVIS. 
In BVISD, we first merge the similar categories into their super-category to avoid ambiguity, and keep the remaining object categories in YTVIS21 and OVIS. For COCO, we use images that contain at least one object in the categories of BVISD. The distribution of the objects in BVISD is illustrated in \cref{fig:bvisd_cate}. 

The statistics of YTVIS21, OVIS, COCO and BVISD are shown in \cref{tab:bvisd_data}. Overall, the proposed BVISD consists of 222k frames, 978k objects and 40 non-conflicting object categories with only object box annotations.
It has been shown \cite{papadopoulos2017extreme,lin2014microsoft} that labeling the bounding box of an object takes only 7 seconds, while labeling its polygon-based mask needs 79.2 second in COCO \cite{lin2014microsoft}. One can estimate that annotating the object bounding boxes in BVISD takes $7 \times 978,000$ seconds (about 80 worker days), while labeling the pixel-wise annotations takes $79.2 \times 978,000$ seconds (about 898 worker days). 

\begin{table}[t]
    \setlength{\tabcolsep}{0.5mm}{
          \linespread{2}
          \begin{tabular}
          {p{0.24\linewidth}p{0.2\linewidth}<{\centering}p{0.15\linewidth}<{\centering}p{0.15\linewidth}<{\centering}p{0.18\linewidth}<{\centering}}
          \Xhline{0.8pt}
            \ Datasets    & \multicolumn{1}{c}{{\color{cyan} YTVIS21}} & \multicolumn{1}{c}{{\color{blue} OVIS}} & \multicolumn{1}{c}{{\color{red} COCO}} &  BVISD \\
            \Xhline{0.8pt}
            \ Videos      & 2985    & 901  & -    & 3886 \\ 
            \ Frames      & 90k     & 42k  & 90k  & 222k \\
            \ Instances   & 8k      & 5k   & 450k & 463k \\
            \ Objects     & 232k    & 296k & 450k & 978k \\
            \ Categories  & 40      & 25   & 80   & 40 \\ 
            \Xhline{0.5pt}
            \ Pixel annot. & 213d    & 272d & 413d & 898d \\
            \ Box annot.   & 19d     & 24d  & 37d  & 80d \\
            \ OL cate.     & 39      & 25   & 25   & 40 \\
            \Xhline{0.8pt}
          \end{tabular}
    }
    \vspace{-2mm}
    \captionof{table}{ Statistics of different VIS datasets. `OL cate.' refers to the number of overlapping categories with BVISD.}
    \label{tab:bvisd_data}
    \vspace{-2mm}
\end{table}

To more easily test the models, we split the BVISD dataset into two separated parts: the training set and the valid set, with the valid set containing around 10\% videos of BVISD. 
As shown in \cref{fig:bvisd_cate}, some categories are only presented in YTVIS21 but not OVIS and COCO, which may result in unbalanced categories. During training, therefore, we sample the clips from the three datasets with different sampling weights to alleviate the issue. 
During inference, we test our proposed BoxVIS with three valid sets: official YTVIS21 valid, official OVIS valid and our BVISD valid. When testing on official YTVIS21 and OVIS valid sets, the predicted categories will be mapped back to the original object categories of the source dataset.
The code to generate BVISD can be found at \url{https://github.com/MinghanLi/BoxVIS}.

\section{Experiments}

\textbf{Datasets.} There are two VIS datasets with pixel-wise annotations: YouTube-VIS series \cite{yang2019video} and OVIS \cite{qi2021occluded}. For YouTube-VIS \cite{yang2019video}, we use the updated 2021 version (YTVIS21) in this paper. YTVIS21 contains 2,985 training, 421 validation, and 453 test videos over 40 categories. All the videos are annotated for every 5 frames. The number of frames per video is between 19 and 36. 
OVIS \cite{qi2021occluded} includes 607 training, 140 validation and 154 test videos, scoping 25 object categories. Different from YTVIS21, videos in OVIS are longer (up to 292 frames) and have more objects with different occlusion levels per frame. 

Our proposed BVISD with only box annotations contains 3524 videos from the above two VIS benchmarks and around 90k short pseudo video clips converted from COCO in the training set, and 362 videos in the validation set. BVISD combines all video types in both YTVIS21 and OVIS, and consists of 40 object categories.

\textbf{Evaluation metrics.} The metrics, including average precision (AP$_{*}$) at different IoU thresholds, average recall (AR$_{*}$) with different object numbers and the mean value of AP (AP), are adopted for VIS model evaluation. OVIS divides instances into slight occlusion (AP$_\text{so}$), moderate occlusion (AP$_\text{mo}$) and heavy occlusion (AP$_\text{ho}$).

\textbf{Implementation details.} Unless otherwise stated, we employ the same hyper-parameters as Mask2Former-VIS \cite{cheng2021mask2former-video}, which is implemented on top of detectron2 \cite{wu2019detectron2}. All VIS models have been pre-trained on COCO image instance segmentation \cite{lin2014microsoft} with pixel-wise annotations. 

The initial learning rate is set to 0.0001, and it decays by a factor of 0.1 at 10k and 12k iterations. The batch size adopts 16 video clips, and each video clip consists of $3$ frames.
During training, we resize the video clips with their shorter edge size being 320 or 512, and keep their longer edge size below 800. During inference, all frames are resized to a shorter edge size of $360$ on YTVIS21 and $480$ on OVIS and BVISD.
We employ the tracking by query embedding matching method proposed in \cite{huang2022minvis} to associate the instances across clips, and adopt the average of its predicted masks in the overlapping clips as the final masks. 

\begin{table}[!t]
\centering
\begin{tabular}{p{0.11\linewidth}<{\centering}p{0.11\linewidth}<{\centering}p{0.19\linewidth}<{\centering}|p{0.16\linewidth}<{\centering}p{0.12\linewidth}<{\centering}}
\Xhline{0.8pt}
 { $L_{proj}$} & {  $L_{pair}$}    & {$L_{pse\text{-}mask}$} & {{\small YTVIS21}} &{{\small OVIS}} \ \\
\Xhline{0.8pt}
$\checkmark$ &               &             & 36.7 & 18.7  \\
$\checkmark$ & $\checkmark$  &             & 39.3 & 22.1  \\  
$\checkmark$ & $\checkmark$  & 0.5         & 40.5 & 23.3 \\
$\checkmark$ & $\checkmark$  & $\epsilon$  & \textbf{41.2} & \textbf{23.9}  \\
\Xhline{0.8pt}
\end{tabular}
\vspace{-2mm}
\caption{Ablation studies on BoxVIS baseline trained on YTVIS21 or OVIS, where `$\epsilon$' refers to the dynamic threshold for selecting high-quality pseudo masks.} \label{tab:abl_baseline}
\vspace{-1mm}
\end{table}

\subsection{Ablation study}
\textbf{Losses in BoxVIS baseline (M2F-VIS-box).} 
We study the effectiveness of the two box-supervised loss items and pseudo-mask loss in \cref{tab:abl_baseline}. 
One can see that BoxVIS baseline trained with only the projection loss $L_{proj}$ performs poorly, while the introduction of the pairwise affinity loss $L_{pair}$ significantly improves the performance by around 3\% AP. 
The pseudo-mask loss $L_{pse\text{-}mask}$ with the high-quality pseudo masks can further improve the performance by 1.2\% AP, which is comparable to its pixel-supervised counterpart M2F-VIS (see \cref{tab:min_boxvis}).  The use of dynamic threshold for selecting high-quality pseudo masks brings another 0.7\% AP increase in performance.

\textbf{Components of the STPA loss.} In \cref{tab:abl_stpa}, we explore the effects of the spatial-temporally paired pixels $E_s$ and $E_t$, the patch correlation $S_{corr}$, and box-center guided shifting in the proposed STPA loss. We only adopt the box-supervised loss and exclude the pseudo-mask loss to better investigate the performance changes.

The pairwise affinity loss $L_{pair}$ employs the spatially paired pixels ($|E_s|=2$) and the color similarity $S_{lab}$, obtaining 41.1\%, 25.5\% and 31.9\% AP on YTVIS21, OVIS and BVISD datasets, respectively. 
When using the default number of spatially paired pixels in BoxInst \cite{tian2020boxinst} ($|E_s|=8$), the performance is nearly the same. 
Introducing the temporally paired pixels $E_t$ into the STPA loss brings 0.7-0.9\% AP improvements on all VIS datasets. 
However, directly adding the patch correlation $S_{corr}$ to the pairwise affinity leads to significant performance drop on OVIS. This is because on object-crowded OVIS videos, patches with high feature correlation may come from different objects, resulting in incorrect pixel-wise supervision.
Fortunately, by introducing the box-center guided shifting, this issue can be alleviated and the STPA loss achieves good performance on all VIS datasets. 
In addition, increasing the number of temporally paired pixels $|E_t|$ can slightly increase the performance. 
Overall, compared with the pairwise affinity loss used in IIS tasks, our STPA loss designed for VIS tasks can bring about 2\% AP improvement on all VIS datasets.

\begin{table}[!t]
\centering
\begin{tabular}{p{0.045\linewidth}<{\centering}p{0.045\linewidth}<{\centering}p{0.05\linewidth}<{\centering}p{0.05\linewidth}<{\centering}p{0.11\linewidth}<{\centering}|p{0.12\linewidth}<{\centering}p{0.08\linewidth}<{\centering}p{0.09\linewidth}<{\centering}}
\Xhline{0.8pt}
 {\small $|E_s|$} & {\small $|E_t|$}  & {\small $S_{lab}$}  & {\small$S_{corr}$} & {\small Shifting} & {\small YTVIS21} &{\small OVIS}  & {\small BVISD\ }\ \\
\hline
2 &   & $\checkmark$ &     &     & 41.1 & 25.5 & 31.9  \\
8 &   & $\checkmark$ &     &     & 41.2 & 25.2 & 32.3 \\
\hline
\hline
2 & 3 & $\checkmark$ &     &     & 41.9 & 26.2 & 32.6 \\  
2 & 3 & $\checkmark$ &  $\checkmark$  &              & 42.3 & 25.0 & 33.3 \\  
2 & 3 & $\checkmark$ &  $\checkmark$  & $\checkmark$ & 42.6 & 26.5 & 33.8 \\
2 & 5 & $\checkmark$ &  $\checkmark$  & $\checkmark$ & \textbf{43.0}  & \textbf{26.9} & \textbf{34.0} \\
\Xhline{0.8pt}
\end{tabular}
\vspace{-2mm}
\caption{Ablation studies on the STPA loss (without using $L_\text{pse-mask}$), where $|\cdot|$ denotes the number of pixels in the set. 
} \label{tab:abl_stpa}
\vspace{-2mm}
\end{table}
\begin{table}[!t]
\centering
\begin{tabular}{p{0.12\linewidth}<{\centering}p{0.1\linewidth}<{\centering}p{0.11\linewidth}<{\centering}|p{0.12\linewidth}<{\centering}p{0.08\linewidth}<{\centering}p{0.11\linewidth}<{\centering}}
\Xhline{0.8pt}
\multicolumn{3}{c|}{ \small Training data}  & \multicolumn{3}{c}{ \small Inference data} \\
{\small YTVIS21} &{\small OVIS}  & {\small COCO}  & {\small YTVIS21} &{\small OVIS}  & {\small BVISD}\ \\
\Xhline{0.8pt}
1 &   &     & 42.8 &  -   & - \\
  & 1 &     & -    & 26.2 & - \\
\hline\hline
1/2 & 1/2 &       & 41.5 & 27.0 & 32.7 \\
2/3 & 1/3 &       & 42.3 & 27.9 & 34.2 \\
1/3 & 1/3 & 1/3   & 42.6 & 28.7 & 34.0 \\
1/4 & 1/2 & 1/4   & 41.9 & 28.7 & 33.5 \\
1/2 & 1/4 & 1/4   & \textbf{43.2} & \textbf{29.0} & \textbf{35.5} \\
\Xhline{0.8pt}
\end{tabular}
\vspace{-2mm}
\caption{Ablation study on BVISD with different sampling weights during training. 
} \label{tab:abl_bvisd}
\vspace{-2mm}
\end{table}

\begin{table*}[!t]
\begin{center}
\setlength{\tabcolsep}{0.5mm}{
      \linespread{2}
      \begin{tabular}
      {p{0.135\textwidth}p{0.14\textwidth}<{\centering}|p{0.055\textwidth}<{\centering}p{0.05\textwidth}<{\centering}p{0.05\textwidth}<{\centering}p{0.05\textwidth}<{\centering}p{0.055\textwidth}<{\centering}|p{0.055\textwidth}<{\centering}p{0.05\textwidth}<{\centering}p{0.05\textwidth}<{\centering}p{0.05\textwidth}<{\centering}p{0.055\textwidth}<{\centering}|p{0.05\textwidth}<{\centering}p{0.06\textwidth}<{\centering}}
         \Xhline{0.8pt}
         \multirow{2}{*}{\ Method} & \multirow{2}{*}{Training data} & \multicolumn{5}{c|}{YTVIS21} & \multicolumn{5}{c|}{OVIS}  &  \multirow{2}{*}{FPS}  & \multirow{2}{*}{Params} \\
          &  &AP &AP$_{50}$ &AP$_{75}$ &AR$_{1}$ &AR$_{10}$ &AP & AP$_{50}$ & AP$_{75}$&AR$_{1}$&AR$_{10}$  &  & \\
         \Xhline{0.8pt}
         \multicolumn{3}{l}{\textit{ \color{red} Pixel-wise annotations}} \\
         \hline
         \ CrossVIS {\small \cite{yang2021crossover}}  & {\small YTVIS21 / OVIS} &  33.3 & 53.8 & 37.0 & 30.1 & 37.6 & 14.9 & 32.7 & 12.1 & 10.3 & 19.8  & 39.8 & 37.5M \\ 
         \ VisTR {\small \cite{wang2020vistr}}        & {\small YTVIS21 / OVIS} &  31.8 & 51.7 & 34.5 & 29.7 & 36.9 & 10.2 & 25.7 & 7.7  & 7.0 & 17.4  & 30.0 & 57.2M \\
         \ IFC  {\small \cite{hwang2021video}}         & {\small YTVIS21 / OVIS} &  36.6 & 57.9 & 39.3 &-     &-     & 13.1 & 27.8 & 11.6 & 9.4 & 23.9  & 46.5 & 39.3M \\
         \ SeqFormer {\small \cite{seqformer}}         & {\small YTVIS21 / OVIS} & 40.5 & 62.4 & 43.7 & 36.1 & 48.1 & 15.1 & 31.9 & 13.8 & 10.4 & 27.1 & 72.3 &  49.3M \\
         \ IDOL {\small \cite{IDOL}}                   & {\small YTVIS21 / OVIS} & 43.9 & \textbf{68.0} & \textbf{49.6} & 38.0 & 50.9 & 24.3 & {45.1} & 23.3 & {14.1} & \underline{33.2} & 30.6  & 43.1M \\
         \ M2F-VIS {\small \cite{cheng2021mask2former-video}} & {\small YTVIS21 / OVIS} & 43.1 & 63.3 & 46.2 & 39.1 & 51.0 & {24.5} & {44.6} & {23.6} & 12.8 & {29.2} & 68.8 & 44.0M \\
         \ MinVIS {\small \cite{huang2022minvis}}  & {\small YTVIS21 / OVIS}  &  {44.2} & 66.0 & 48.1 &\underline{39.2}  &\underline{51.7}  & \underline{26.3} & \underline{47.9} & \underline{25.1} & \textbf{14.6} & 30.0 & 52.4 & 44.0M \\
         \ VITA {\small \cite{heo2022vita}}            & {\small YTVIS21 / OVIS} & \textbf{45.7} & \underline{67.4} & \underline{49.5} & \textbf{40.9} & \textbf{53.6} & 19.6 & 41.2 & 17.4 & 11.7 & 26.0  & 33.7 & 57.2M \\
         \ MDQE {\small \cite{li2023mdqe}} & {\small YTVIS21 / OVIS} & \underline{44.5} & 67.1 & 48.7 & 37.9 & 49.8 & \textbf{29.2} & \textbf{55.2} & \textbf{27.1} & \underline{14.5} & \textbf{34.2} & 37.8 & 51.4M \\
         \Xhline{0.8pt}
         \multicolumn{3}{l}{\textit{\color{red} Box annotations}} \\
         \hline
         \ M2F-VIS-box  & {\small YTVIS21 / OVIS} &  41.2 & 63.3 & 44.6 & 37.9 & 48.5 & 23.9 & 46.5 & 22.8 & 13.6 & 28.1 & 37.8 & 44.0M \\ 
         \ BoxVIS (our) & {\small YTVIS21 / OVIS} & 42.8 & 64.9 & \textbf{47.3} & 38.0 & 49.3 & 26.2 & 49.4 & 26.3 & 13.6 & 30.0 & 37.8 & 44.0M \\
         \ BoxVIS (our) & \quad BVISD &  \textbf{43.2} & \textbf{66.8} & {46.6} & \textbf{39.0} & \textbf{50.7} & \textbf{29.0} & \textbf{52.2} & \textbf{29.1} & \textbf{14.8} & \textbf{33.1} & 37.8 & 44.0M \\
        \Xhline{0.8pt}
      \end{tabular}
}
\end{center}
\vspace{-5mm}
\caption{ Quantitative performance comparison of pixel-supervised and box-supervised VIS methods with ResNet50 backbone on YTVIS21 and OVIS, where FPS is computed on YTVIS21 valid set. Best in \textbf{bold}, and second best with \underline{underline}.
}\label{tab:sota_yt21_ovis}
\vspace{-1mm}
\end{table*}

\textbf{Sampling weights in BVISD.} Due to the unbalanced number of objects per category in BVISD (see \cref{fig:bvisd_cate}), we propose to sample the video clips from the three datasets with different sampling weights during training. Taking the last row of \cref{tab:abl_bvisd} as an example, among the 8 short video clips, 4, 2 and 2 are from YTVIS21, OVIS and COCO, respectively.
BoxVIS trained {\it separately} on YTVIS21 or OVIS obtains 42.8\% and 26.2\% AP, respectively, while the model trained jointly on YTVIS21 and OVIS with equal sampling weights shows a performance drop on YTVIS21 but a slight performance increase on OVIS. 
This is because some object categories presented in YTVIS21 are not shown in OVIS, and the object-crowded videos in OVIS contribute more to gradient back-propagation in training. By setting a higher sampling weight on YTVIS21, the performance is much improved on all VIS datasets.
By introducing the pseudo video clips from COCO and setting higher sampling weights on YTVIS21 and COCO, higher performance can be achieved. In addition, the generated pseudo video clips from COCO can bring about 1\% AP improvement.
Unless specified, the sampling weights in BVISD are set as $1/2, 1/4$ and $1/4$ for YTVIS21, OVIS and COCO, respectively.

\begin{table*}[!t]
\centering
\begin{subtable}{0.5\textwidth}
\setlength{\tabcolsep}{0.5mm}{
      \linespread{2}
      \begin{tabular}
      {p{0.255\linewidth}p{0.11\linewidth}<{\centering}p{0.1\linewidth}<{\centering}p{0.1\linewidth}<{\centering}p{0.1\linewidth}<{\centering}p{0.1\linewidth}<{\centering}p{0.11\linewidth}<{\centering}}
        \Xhline{0.8pt}
         Method & Super. &\ AP\  &\ AP$_{50}$\ &\ AP$_{75}$\ &\ AR$_{1}$\ &\ AR$_{10}$\ \\
         \Xhline{0.8pt}
         {\normalsize M2F-VIS} \cite{cheng2021mask2former-video} & Pixel  & 52.6 & 76.4 & 57.2 & - & -\\
         {\normalsize MinVIS} \cite{huang2022minvis}   & Pixel & 55.3 & 76.6 & \underline{62.0} & \underline{45.9} & \underline{60.8} \\
         {\normalsize VITA} \cite{heo2022vita}         & Pixel & \textbf{57.5} & \underline{80.6} & 61.0 & \textbf{47.7} & \textbf{62.6} \\
         {\normalsize IDOL} \cite{IDOL}                & Pixel & 56.1 & \textbf{80.8} & \textbf{63.5} & 45.0  &60.1 \\
         MDQE \cite{li2023mdqe}                        & Pixel & \underline{56.2} & 80.0 & 61.1 & 44.9 & 59.1 \\
         \hline
         {\normalsize BoxVIS}                          & Box   & 52.8 & 75.4 & 58.3 & 44.4 & 58.1 \\
         {\normalsize BoxVIS$^*$}                      & Box   & 53.9 & 76.4 & 59.6 & 44.8 & 61.0 \\
         \Xhline{0.8pt}
      \end{tabular}
}
\caption{Performance comparison on YTVIS21}\label{tab:sota_swinl_yt}
\end{subtable}
\begin{subtable}{0.49\textwidth}
\setlength{\tabcolsep}{0.5mm}{
      \linespread{2}
      \begin{tabular}
      {p{0.23\textwidth}p{0.11\textwidth}<{\centering}p{0.09\textwidth}<{\centering}p{0.09\textwidth}<{\centering}p{0.09\textwidth}<{\centering}p{0.09\textwidth}<{\centering}p{0.09\textwidth}<{\centering}p{0.1\textwidth}<{\centering}}
        \Xhline{0.8pt}
          Method &{\normalsize Super.} & { AP} & {AP$_{50}$} & { AP$_{75}$} & { AP$_{so}$} & { AP$_{mo}$} &  { AP$_{ho}$}\ \\
         \Xhline{0.8pt}
         M2F-VIS \cite{cheng2021mask2former-video} & Pixel & 26.4 & 50.2 & 26.9 & 46.1 & 30.9 & 9.5 \\ 
         MinVIS \cite{huang2022minvis}  & Pixel   & \underline{41.6} & 65.4 & 43.4 & \textbf{65.5} & \underline{48.5} & \underline{20.5} \\ 
         VITA \cite{heo2022vita}        & Pixel  & 27.7 & 51.9 & 24.9 &  - & - & - \\ 
         IDOL$^\dag$\cite{IDOL}         & Pixel  & \textbf{42.6} & \underline{65.7} & \textbf{45.2} &  - & - & -  \\ 
         MDQE$^\dag$ \cite{li2023mdqe}  & Pixel  & \textbf{42.6} & \textbf{67.8} & \underline{44.3} & \underline{65.1} & \textbf{49.3} & \textbf{21.6} \\ 
         \hline
         BoxVIS                         & Box    & 34.4 & 59.0 & 33.8 & 56.8 & 39.5 & 14.6 \\
         BoxVIS$^*$                     & Box    & 40.6 & 68.4 & 39.9 & 59.4 & 45.8 & 20.9 \\
         \Xhline{0.8pt}
      \end{tabular}
}
\caption{Performance comparison on OVIS}\label{tab:sota_swinl_ovis}
\end{subtable}
\vspace{-1.5mm}
\caption{ Quantitative performance comparison of pixel-supervised and box-supervised VIS methods with Swin Large (SwinL) backbone \cite{vaswani2017attention} on YTVIS21 and OVIS, respectively. Symbol `$\dag$'  means that the input video is of 720p. BoxVIS with symbol `$*$' indicates that the results are obtained by using the clip-level tracker proposed in MDQE \cite{li2023mdqe}.
}\label{tab:sota_swinl}
\end{table*}

\begin{figure*}
    \centering
    \includegraphics[width=0.9\textwidth]{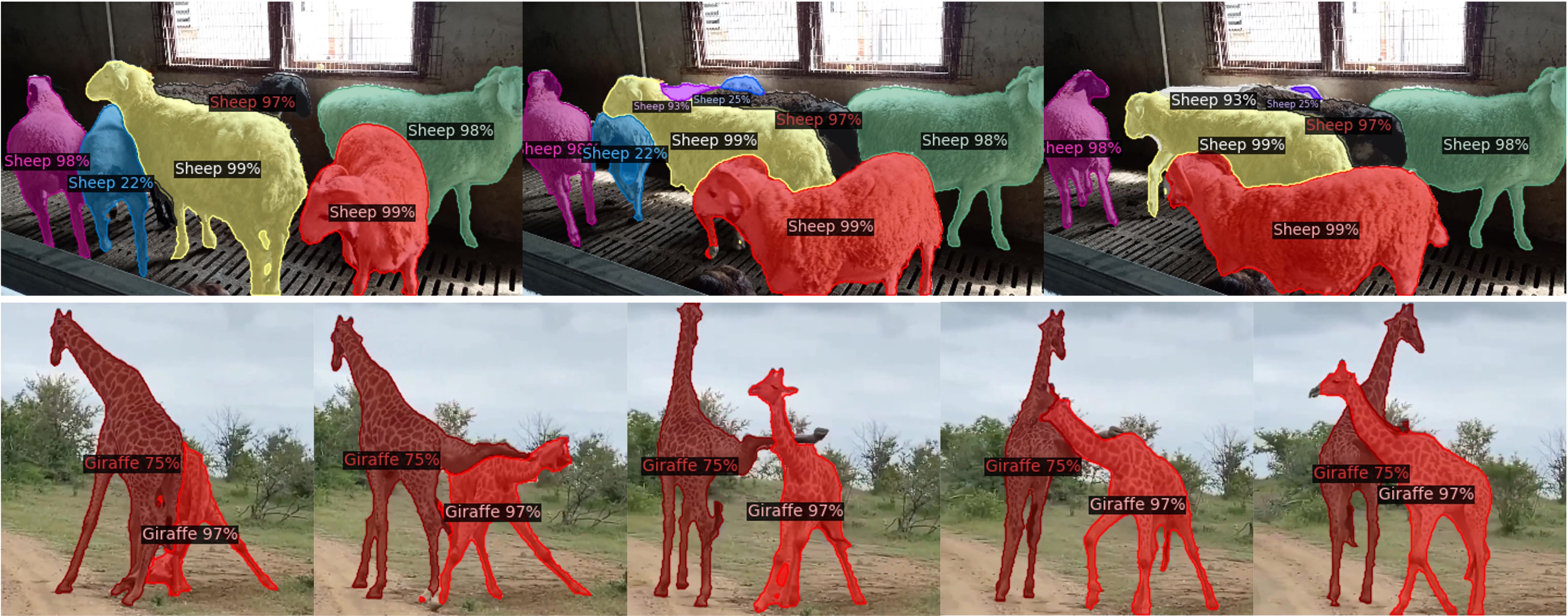}
    \vspace{-2mm}
\caption{Box-supervised instance segmentation on two challenging videos in OVIS valid set. The first row shows the results of crowded objects with heavy occlusion, while the second row shows the results of two objects with complex boundary. }
\label{fig:visualizer}
\vspace{-2mm}
\end{figure*}
\subsection{Main results}
With ResNet50 backbone and Swin Large backbone, we compare the proposed BoxVIS with the state-of-the-art pixel-supervised VIS methods on YTVIS21 and OVIS datasets in \cref{tab:sota_yt21_ovis} and \cref{tab:sota_swinl}, respectively.

\textbf{YTVIS21 valid set.} 
From \cref{tab:sota_yt21_ovis}, we can see that the accuracy of early pixel-supervised VIS methods \cite{yang2021crossover, wang2020vistr, hwang2021video, seqformer} is below 41\% AP.
The recently developed pixel-supervised VIS methods with self-supervised learning or masked-attention transformer bring about 4\% and 5\% AP improvement. Trained jointly on VIS and COCO datasets, VITA \cite{heo2022vita} and MDQE \cite{li2023mdqe} achieve 45.7\% and 44.5\% AP, respectively.
M2F-VIS-box (our BoxVIS baseline) obtains 41.2\% AP, 1.9\% lower than its pixel-supervised counterpart M2F-VIS \cite{cheng2021mask2former-video}. In contrast, our BoxVIS with the STPA loss can increase the performance to 42.8\% AP, and our BoxVIS trained on BVISD improves the performance to 43.2\% AP. Overall, our BoxVIS outperforms its pixel-supervised counterpart M2F-VIS \cite{cheng2021mask2former-video} and exhibits comparable performance to the state-of-the-art pixel-supervised competitors.

\textbf{OVIS valid set.} As shown in \cref{tab:sota_yt21_ovis}, the early proposed pixel-supervised VIS methods fail to handle the challenging videos in OVIS dataset, resulting in poor performance (below 15\% AP). The recent masked-attention based methods M2F-VIS \cite{cheng2021mask2former-video} and MinVIS \cite{huang2022minvis} bring approximately 10\% AP improvement, achieving 24.5\% and 26.3\% AP, respectively. By introducing clip-level input and contrastive learning for object embeddings, MDQE \cite{li2023mdqe} obtains the state-of-the-art 29.2\% AP.
However, VITA \cite{heo2022vita}, which is the state-of-the-art on YTVIS21, obtains only 19.6\% AP on OVIS, showing limited generalization capability.
The box-supervised baseline M2F-VIS-box and our BoxVIS trained only on OVIS obtain 23.9\% and 26.2\% AP, respectively, validating that the proposed STPA loss can predict instance masks with better spatial and temporal consistency. Finally, our BoxVIS trained on BVISD improves the performance to 29.0\% AP, obtaining competitive performance with the state-of-the-art pixel-supervised VIS competitors. The encouraging results demonstrate great potentials of box annotation in VIS tasks.

\textbf{Swin Large backbone.} VIS models with the stronger Swin Large backbone can have higher detection and segmentation abilities. Due to the limited space, only the recently proposed high-performance methods are reported in \cref{tab:sota_swinl}. 
Among the pixel-supervised VIS methods, VITA \cite{heo2022vita} shows the best accuracy (57.5\% AP) on YTVIS21, while MinVIS \cite{huang2022minvis}, IDOL \cite{IDOL} and MDQE \cite{li2023mdqe} achieve 41.6\%, 42.6\% and 42.6\% AP, respectively, where IDOL \cite{IDOL} and MDQE \cite{li2023mdqe} take 720p videos as input.

Our BoxVIS with the frame-level tracker \cite{huang2022minvis} achieves 52.8\% and 34.4\% AP on YTVIS21 and OVIS, respectively, outperforming its pixel-supervised counterpart M2F-VIS \cite{cheng2021mask2former-video}.
However, there is a performance gap on OVIS dataset compared to the top-performance pixel-supervised VIS methods. We believe one reason is that for challenging videos on OVIS, the transformer network with attention mechanism can learn more discriminative object embeddings from pixel-wise annotation than box-level annotation.
Fortunately, since our BoxVIS follows the clip-level input pipeline, it can introduce the mask IoU on overlapping frames into the tracker to more accurately associate objects across clips. In particular, by using the clip-level tracker proposed in MDQE \cite{li2023mdqe}, the performance of BoxVIS can be significantly improved to 53.9\% and 40.6\% AP on YTVIS21 and OVIS, respectively, which is competitive with the pixel-supervised VIS methods.
It is worth mentioning that our BVISD consumes only 16\% the cost of pixel-wise annotation, and there is still a big room for BoxVIS to improve its performance.

\textbf{Visualization.} \cref{fig:visualizer} displays the predicted instance masks by BoxVIS on OVIS videos. One can see that with only box annotations, BoxVIS can still predict high-quality instance masks for challenging videos with occluded objects and complex boundaries. Visualization comparison on more videos are provided in the \textbf{supplemental materials}. 

\textbf{Parameters and Speed.} We follow Detectron2 \cite{wu2019detectron2} to calculate the number of parameters and FPS for all VIS methods in \cref{tab:sota_yt21_ovis}. BoxVIS inherits the network architecture of M2F-VIS \cite{cheng2021mask2former-video} and thus has the same model size (44M) as it. The tracking strategy affects mostly the inference speed of VIS models.
SeqFormer\cite{seqformer} and M2F-VIS \cite{cheng2021mask2former-video} take the whole video as input without extra tracking, running at around 70FPS. IDOL \cite{IDOL} and MinVIS \cite{huang2022minvis} with frame-by-frame inference run at 30.6FPS and 52.4FPS, respectively. 
Our BoxVIS adopts clip-by-clip inference to obtain averaged masks on the overlapping clips, resulting in a slower inference speed of 37.8FPS. If we set the clip length as 1, BoxVIS can run as fast as M2F-VIS (\ie, 70FPS) at the price of about 1\% AP drop on all VIS datasets.

\section{Conclusion}
We explored the feasibility of using only box annotations for VIS. We first extended the pixel-supervised VIS methods to a box-supervised VIS (BoxVIS) baseline, and proposed a spatial-temporal pairwise affinity (STPA) loss to introduce temporal pixel-wise mask supervision. Finally, we collected a large scale box-supervised VIS dataset (BVISD). The experimental results showed that the proposed BoxVIS trained on BVISD can yield high-quality object masks with decent spatial and temporal consistency, achieving comparable or even better performance than pixel-supervised VIS methods on YTVIS21 and OVIS. It is a promising direction to further investigate more effective BoxVIS models. 

\newpage
{\small
\bibliographystyle{ieee_fullname}
\bibliography{paper_for_review}
}

\newpage
\onecolumn
\vspace{0.2cm}
{\centering\section*{\Large Supplementary Materials}}
\vspace{0.5cm}

In this supplementary file, we provide the following materials:
\begin{itemize}
    \item An illustrative example of pseudo clip generation ($cf.$ Section 3.3 in the main paper);
    \item More visual comparisons of pixel-supervised and box-supervised VIS methods ($cf.$ Section 4.2 in the main paper);
    \item More visualizations of instance masks predicted by BoxVIS ($cf.$ Section 4.2 in the main paper).
\end{itemize}

\noindent\textbf{A. An illustrative example of pseudo clip generation}

\begin{figure}[h]
    \centering
    \vspace{-1mm}
    \includegraphics[width=0.6\linewidth]{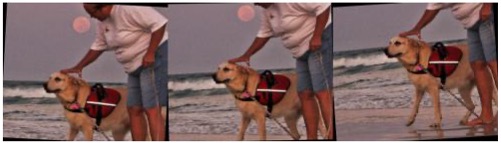}
    \vspace{-2mm}
    \caption{An illustrated example of pseudo clip generation.}
    \label{fig:coco2clip}
    \vspace{-1mm}
\end{figure}

\cref{fig:coco2clip} shows an example on how to convert a static image from the COCO dataset to a pseudo 3-frame video clip. The augmentation process consists of three geometric transformations: zoom, crop and rotate images.

\noindent\textbf{B. More visual comparisons of pixel-supervised and box-supervised VIS methods}

In Figs. { \color{red} 5} - { \color{red} 7}, we provide more visual comparisons of instance masks predicted by M2F-VIS \cite{cheng2021mask2former-video}, M2F-VIS-box and our BoxVIS on videos with crowded scenes, blurred or occluded objects. It can bee seen that compared to the box-supervised baseline M2F-VIS-box, our BoxVIS can predicate instance masks with better spatial and temporal consistency. On the other hand, BoxVIS can yield instance masks of the same quality as the pixel-supervised M2F-VIS \cite{cheng2021mask2former-video}.

\

\noindent\textbf{C. Visualization of instance masks predicted by BoxVIS} 

\begin{figure*}[h]
    \centering
    \includegraphics[width=0.98\textwidth]{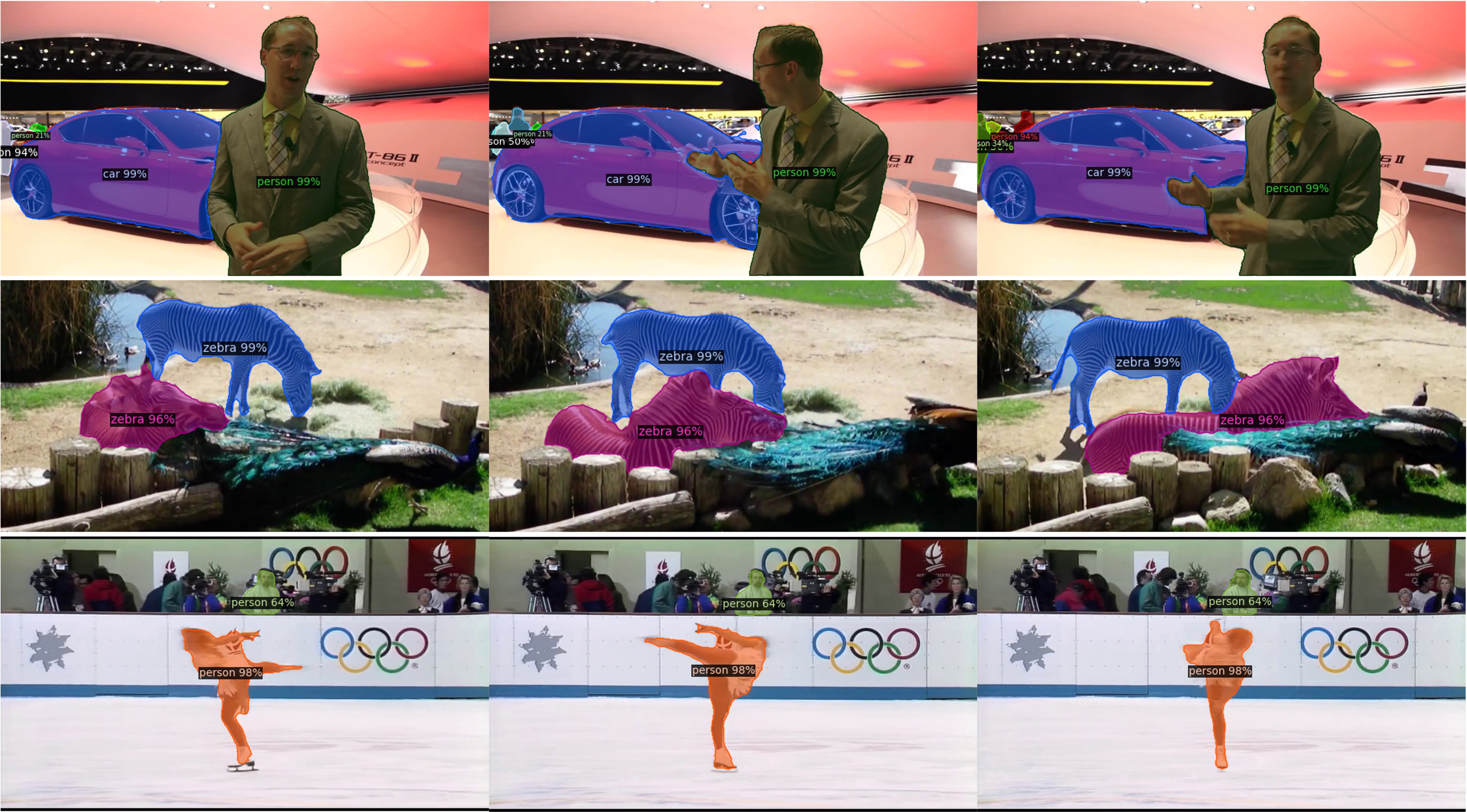}
    \includegraphics[width=0.98\textwidth]{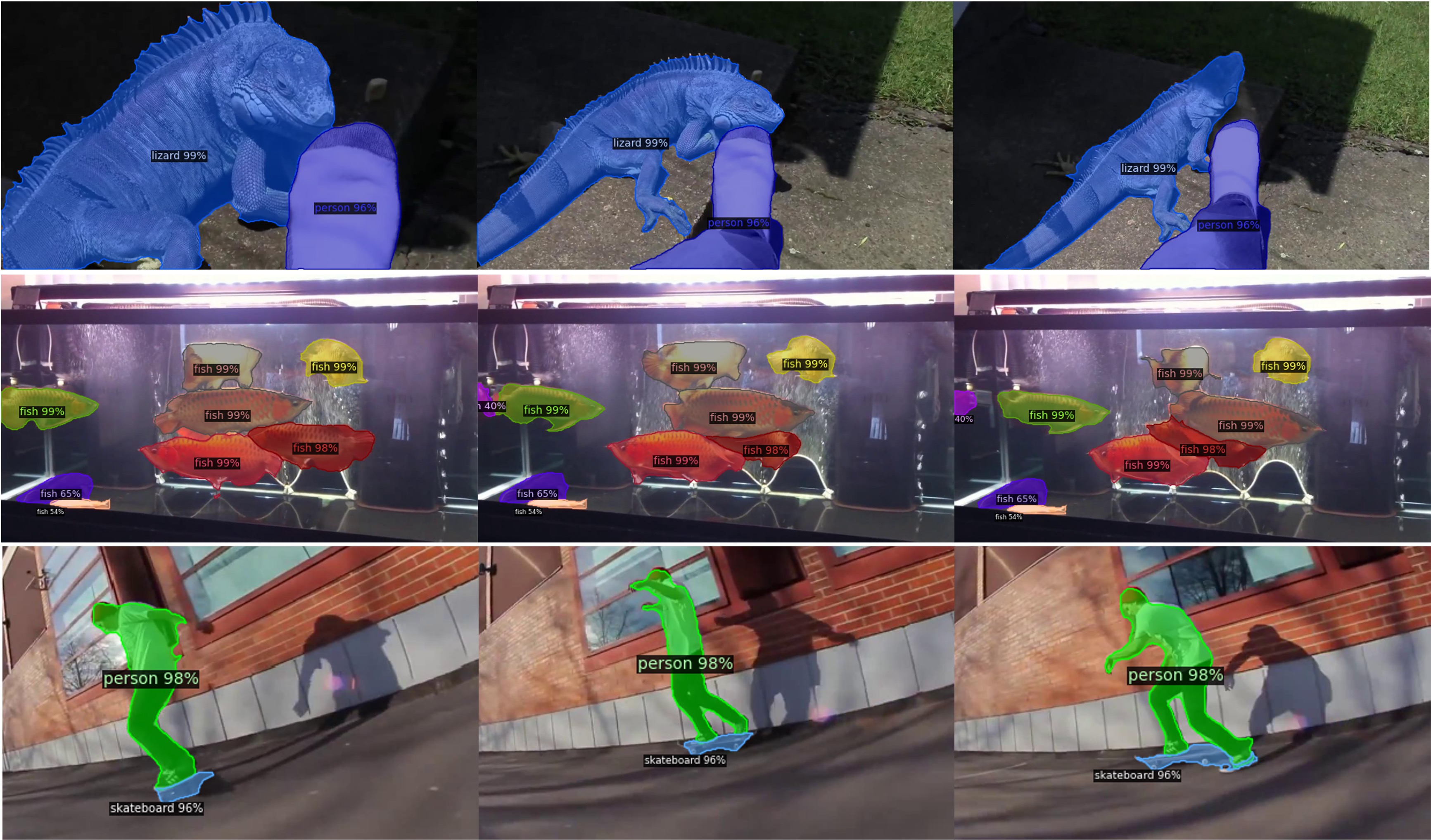}
\caption{Visualization of instance masks predicted by BoxVIS on the YTVIS21 valid set.  }
\label{fig:visualizer_yt21}
\end{figure*}

\begin{figure*}[t]
    \centering
    \includegraphics[width=0.98\textwidth]{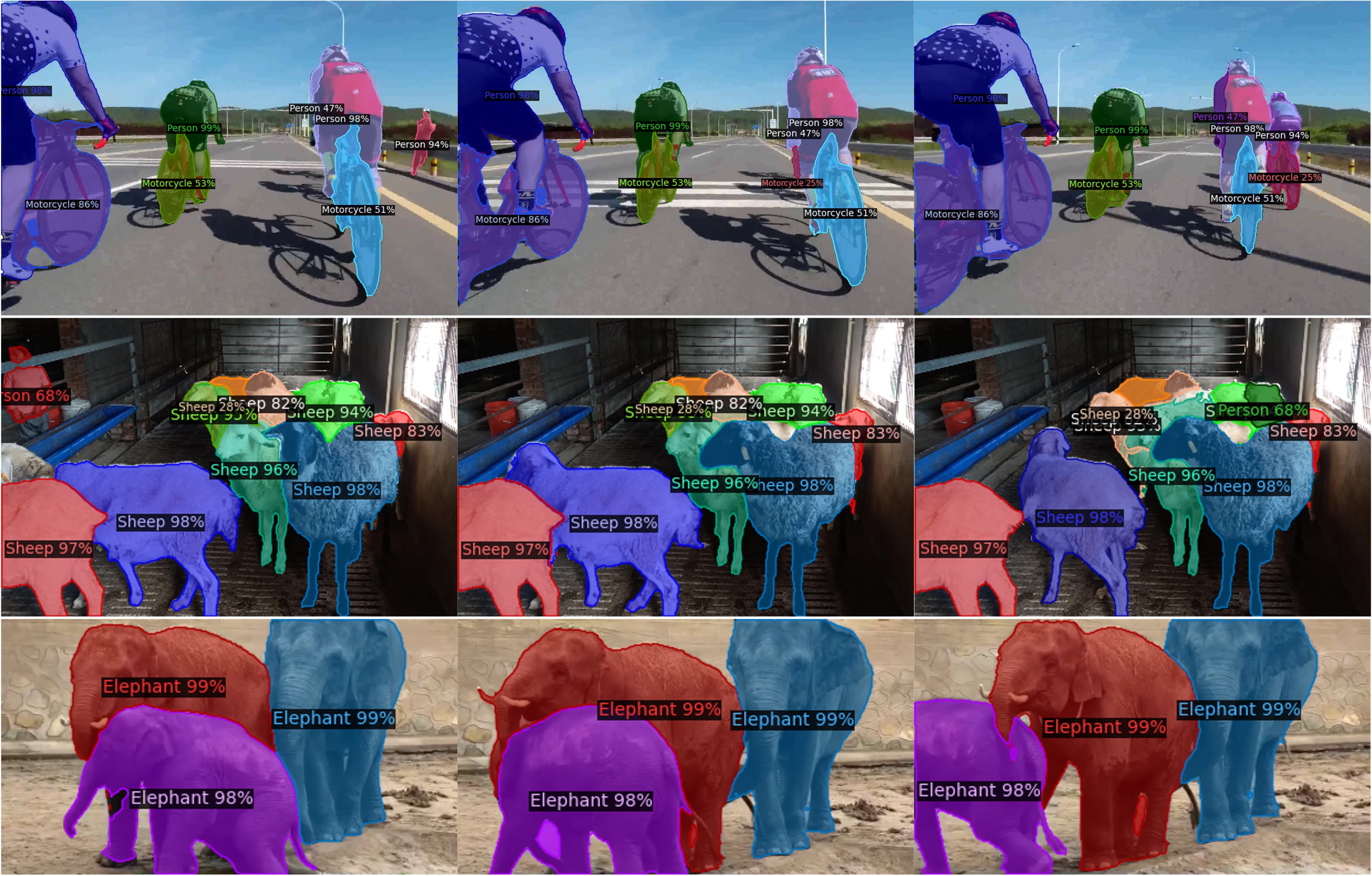}
    \includegraphics[width=0.98\textwidth]{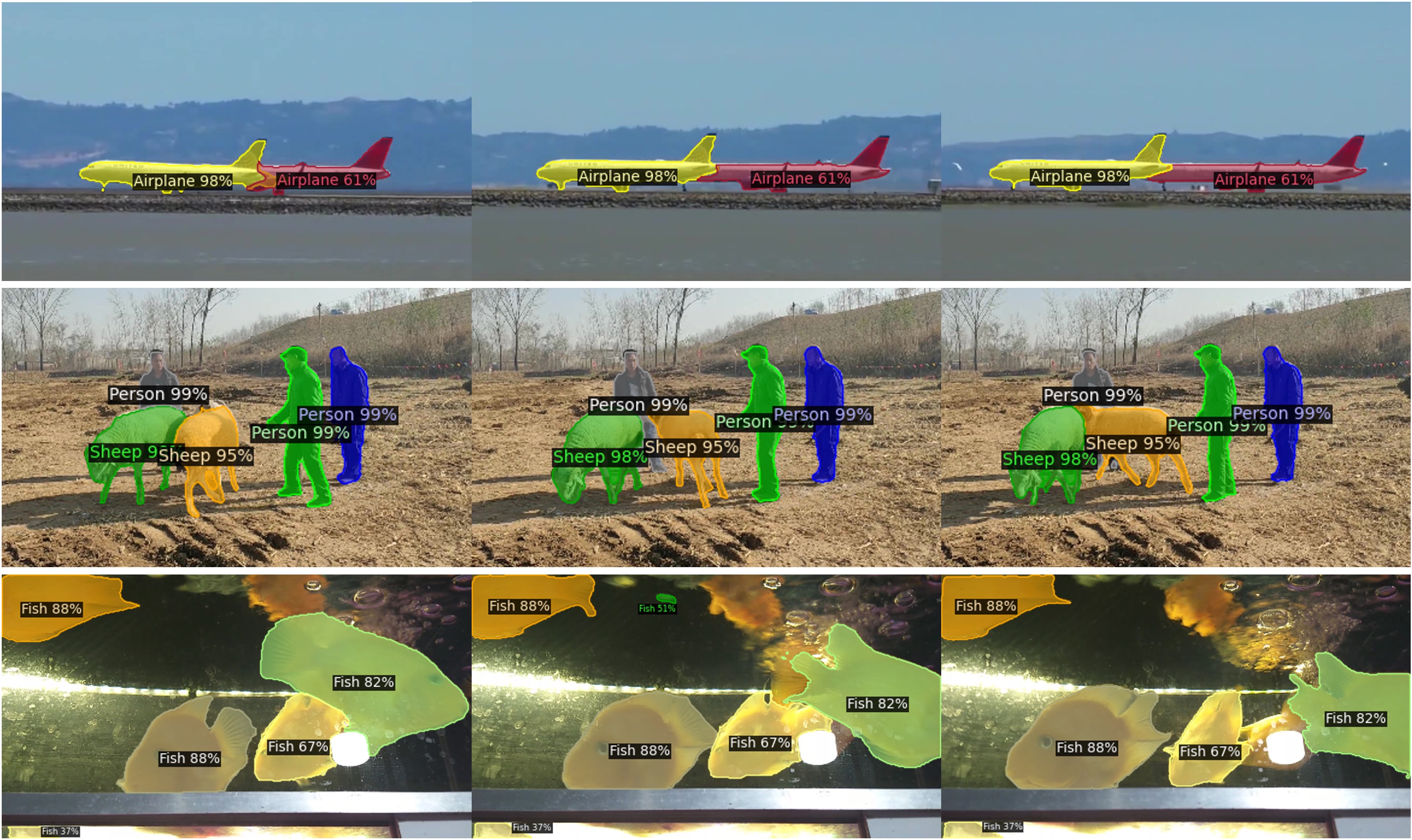}
\caption{Visualization of instance masks predicted by BoxVIS on the OVIS valid set.  }
\label{fig:visualizer_ovis}
\end{figure*}

In \cref{fig:visualizer_yt21} and \cref{fig:visualizer_ovis}, we visualize instance masks predicted by BoxVIS on various videos of YTVIS21 and OVIS, demonstrating the model generalization ability.

\begin{figure*}[h]
    \centering
    \includegraphics[width=0.98\textwidth]{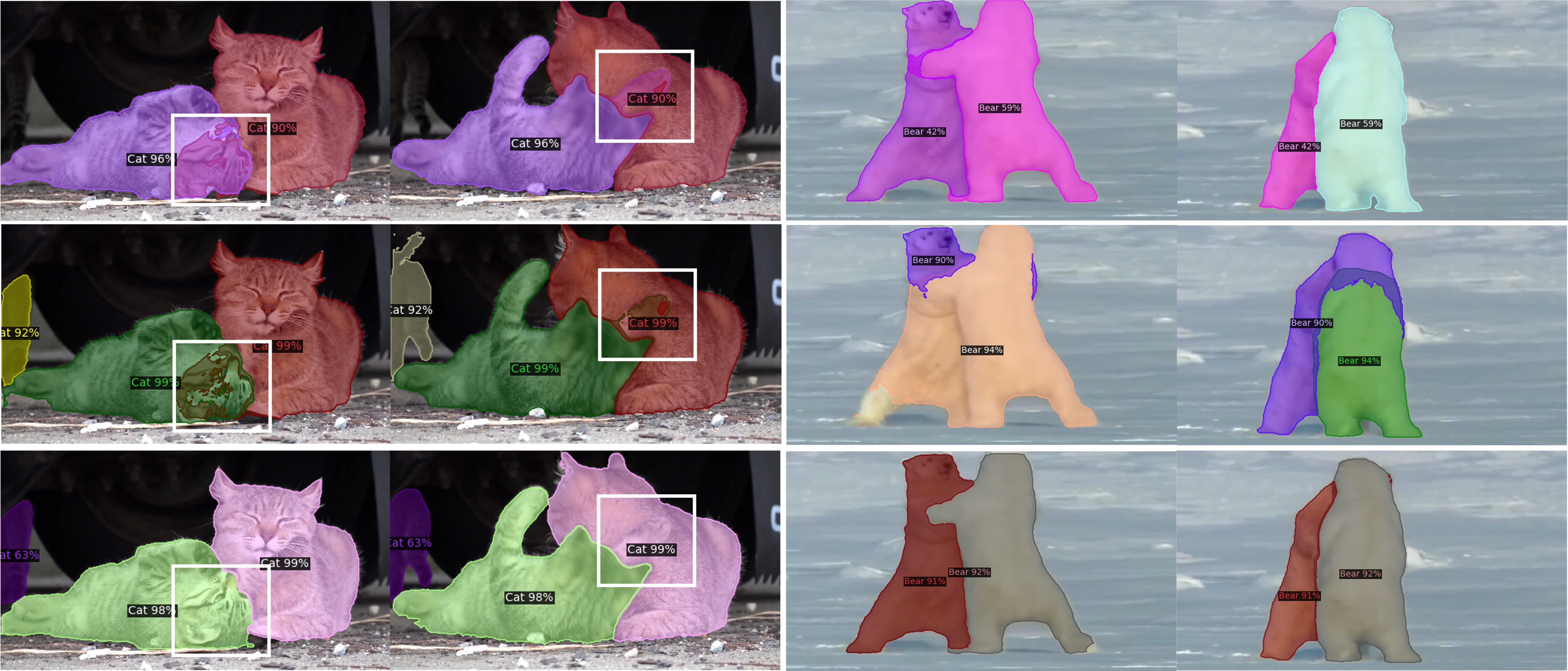}
\caption{Visual comparison of instance segmentation results by M2F-VIS (top), M2F-VIS-box (middle row) and BoxVIS (bottom row) on videos with occluded objects.}
\label{fig:visualizer_occ}
\end{figure*}

\begin{figure*}[h]
    \centering
    \includegraphics[width=0.98\textwidth]{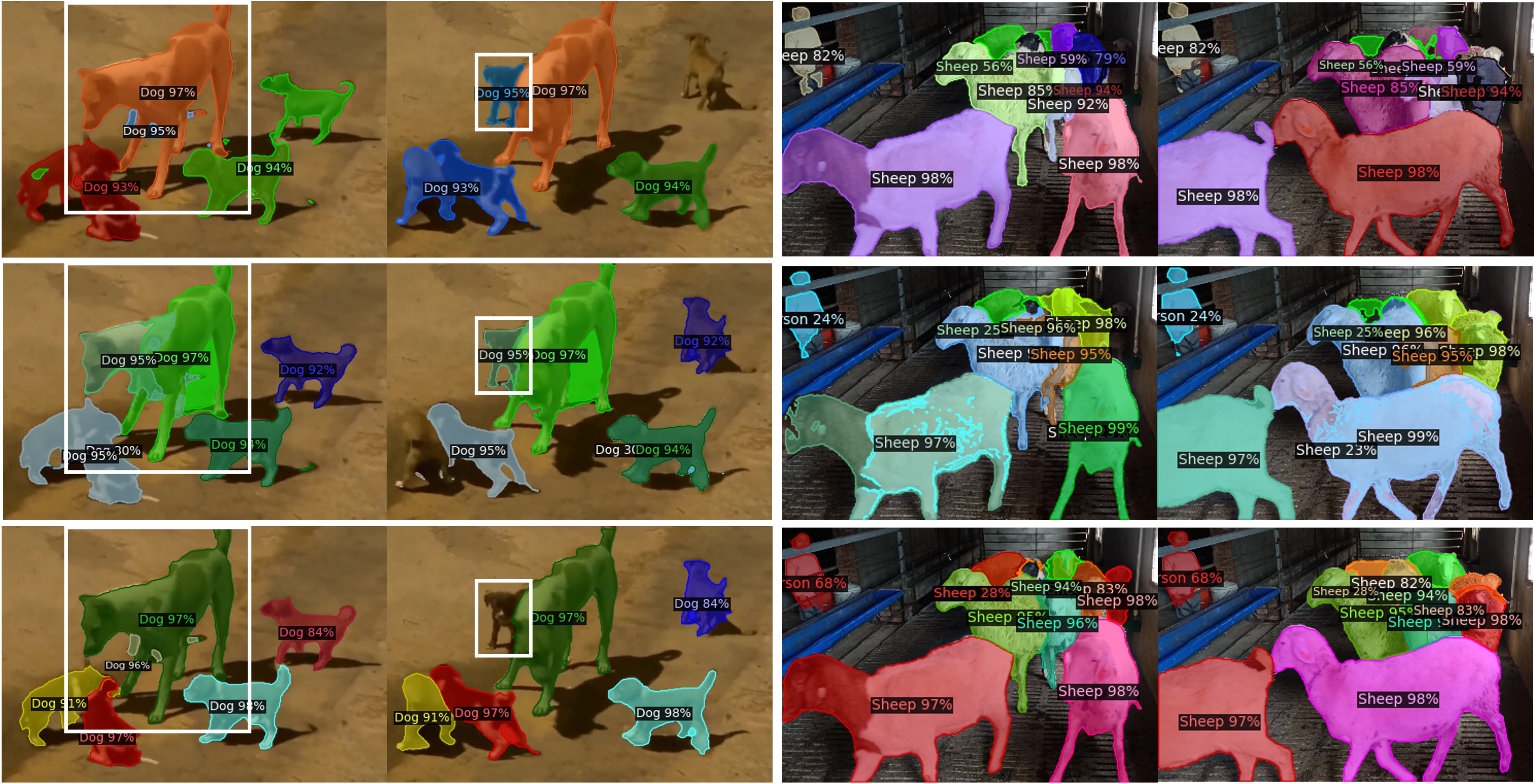}
\caption{Visual comparison of instance segmentation results by M2F-VIS (top), M2F-VIS-box (middle row) and BoxVIS (bottom row) on videos with crowded scenes.}
\label{fig:visualizer_crowded}
\end{figure*}

\begin{figure*}[h]
    \centering
    \includegraphics[width=0.98\textwidth]{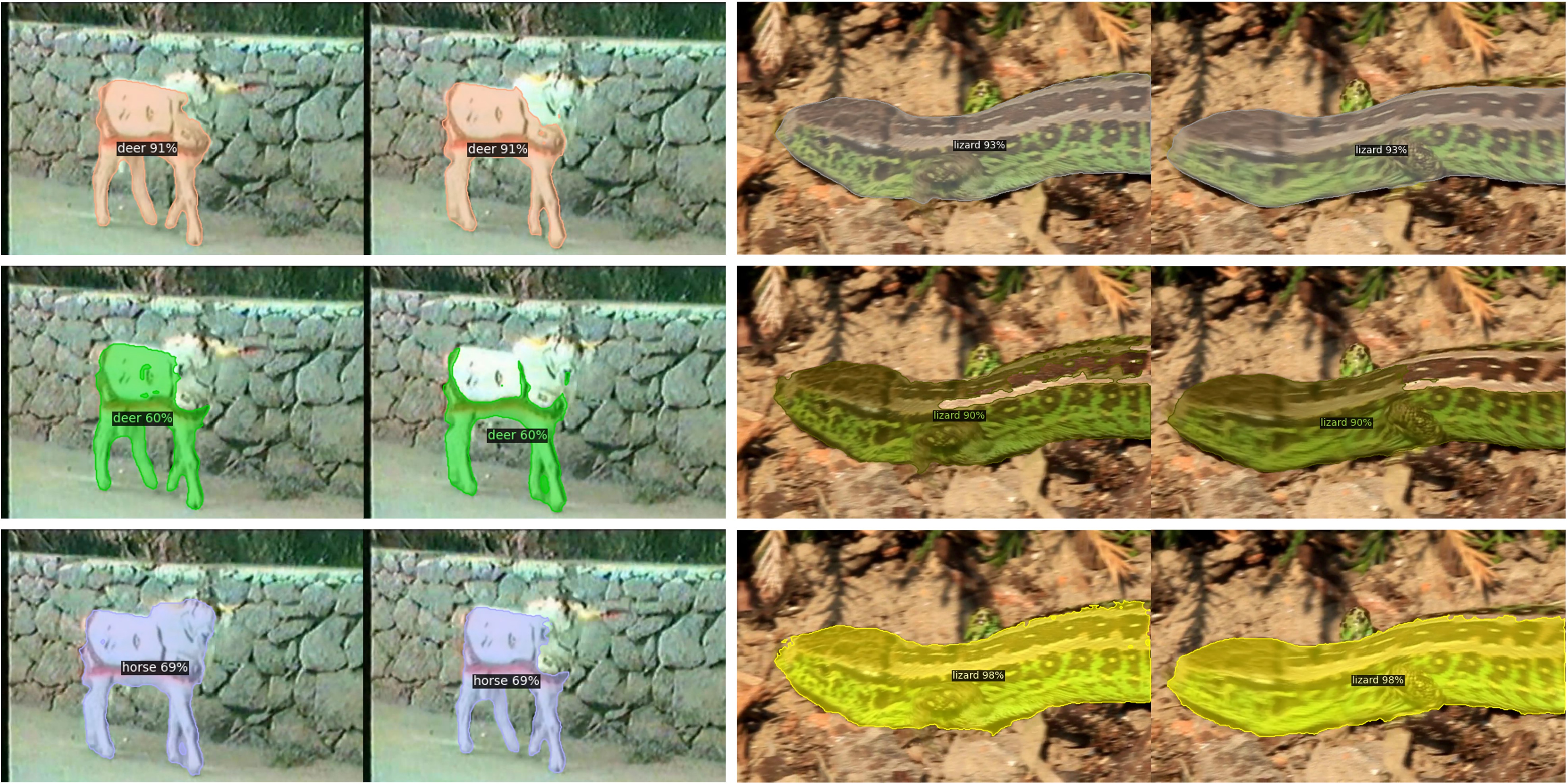}
\caption{Visual comparison of instance segmentation results by M2F-VIS (top), M2F-VIS-box (middle row) and BoxVIS (bottom row) on videos with blurred objects.}
\label{fig:visualizer_motion}
\end{figure*}

\end{document}